\documentclass[letterpaper]{article} 
\usepackage{aaai2026}  
\usepackage{times}  
\usepackage{helvet}  
\usepackage{courier}  
\usepackage[hyphens]{url}  
\usepackage{graphicx} 
\usepackage{natbib}  
\usepackage{caption} 
\usepackage{algorithm}
\usepackage{algorithmic}
\usepackage{amsmath}
\usepackage{amssymb}
\usepackage{multirow}
\usepackage{subcaption}
\usepackage{makecell}
\usepackage{booktabs}
\usepackage{xcolor}
\usepackage{pifont}
\usepackage{newfloat}
\usepackage{listings}
\DeclareCaptionStyle{ruled}{labelfont=normalfont,labelsep=colon,strut=off} 
\floatstyle{ruled}
\newfloat{listing}{tb}{lst}{}
\floatname{listing}{Listing}
\newcounter{checksubsection}
\newcounter{checkitem}[checksubsection]

\usepackage{bibentry}

\title{RS2-SAM2: Customized SAM2 for Referring Remote Sensing Image Segmentation}

\author{
    Fu Rong\textsuperscript{\rm 1}, Meng Lan\textsuperscript{\rm 2},
    Qian Zhang\textsuperscript{\rm 3}, Lefei Zhang\textsuperscript{\rm 1,}\thanks{Corresponding author.}
}

\affiliations{
    \textsuperscript{\rm 1}National Engineering Research Center for Multimedia Software, School of Computer Science, Wuhan University \\
    \textsuperscript{\rm 2}Hong Kong University of Science and Technology \\ 
    \textsuperscript{\rm 3}Horizon Robotics
    \\
    \{furong, zhanglefei\}@whu.edu.cn, eemenglan@ust.hk, qian01.zhang@horizon.auto
}
\begin{document}
\maketitle
\begin{abstract}

Referring Remote Sensing Image Segmentation (RRSIS) aims to segment target objects in remote sensing (RS) images based on textual descriptions. Although Segment Anything Model 2 (SAM2) has shown remarkable performance in various segmentation tasks, its application to RRSIS presents several challenges, including understanding the text-described RS scenes and generating effective prompts from text. To address these issues, we propose \textbf{RS2-SAM2}, a novel framework that adapts SAM2 to RRSIS by aligning the adapted RS features and textual features while providing pseudo-mask-based dense prompts. Specifically, we employ a union encoder to jointly encode the visual and textual inputs, generating aligned visual and text embeddings as well as multimodal class tokens. A bidirectional hierarchical fusion module is introduced to adapt SAM2 to RS scenes and align adapted visual features with the visually enhanced text embeddings, improving the model's interpretation of text-described RS scenes. To provide precise target cues for SAM2, we design a mask prompt generator, which takes the visual embeddings and class tokens as input and produces a pseudo-mask as the dense prompt of SAM2. Experimental results on several RRSIS benchmarks demonstrate that RS2-SAM2 achieves state-of-the-art performance.
\end{abstract}

 \begin{links}
 \link{Code}{https://github.com/rongfu-dsb/RS2-SAM2} 
 \end{links}

\section{Introduction}
Referring Remote Sensing Image Segmentation (RRSIS) \cite{liu2024rotated, yuan2024rrsis} aims to segment specified targets from aerial images based on textual descriptions. This task extends the capabilities of traditional Referring Image Segmentation (RIS) \cite{wang2022cris, yang2022lavt, liu2023polyformer} by addressing the unique challenges inherent to remote sensing images. These challenges include handling diverse spatial scales, interpreting complex scene contexts, and resolving ambiguous object boundaries, which are particularly prevalent in remote sensing scenes.

\begin{figure}[t]
    \centering
     \includegraphics[width=\linewidth]{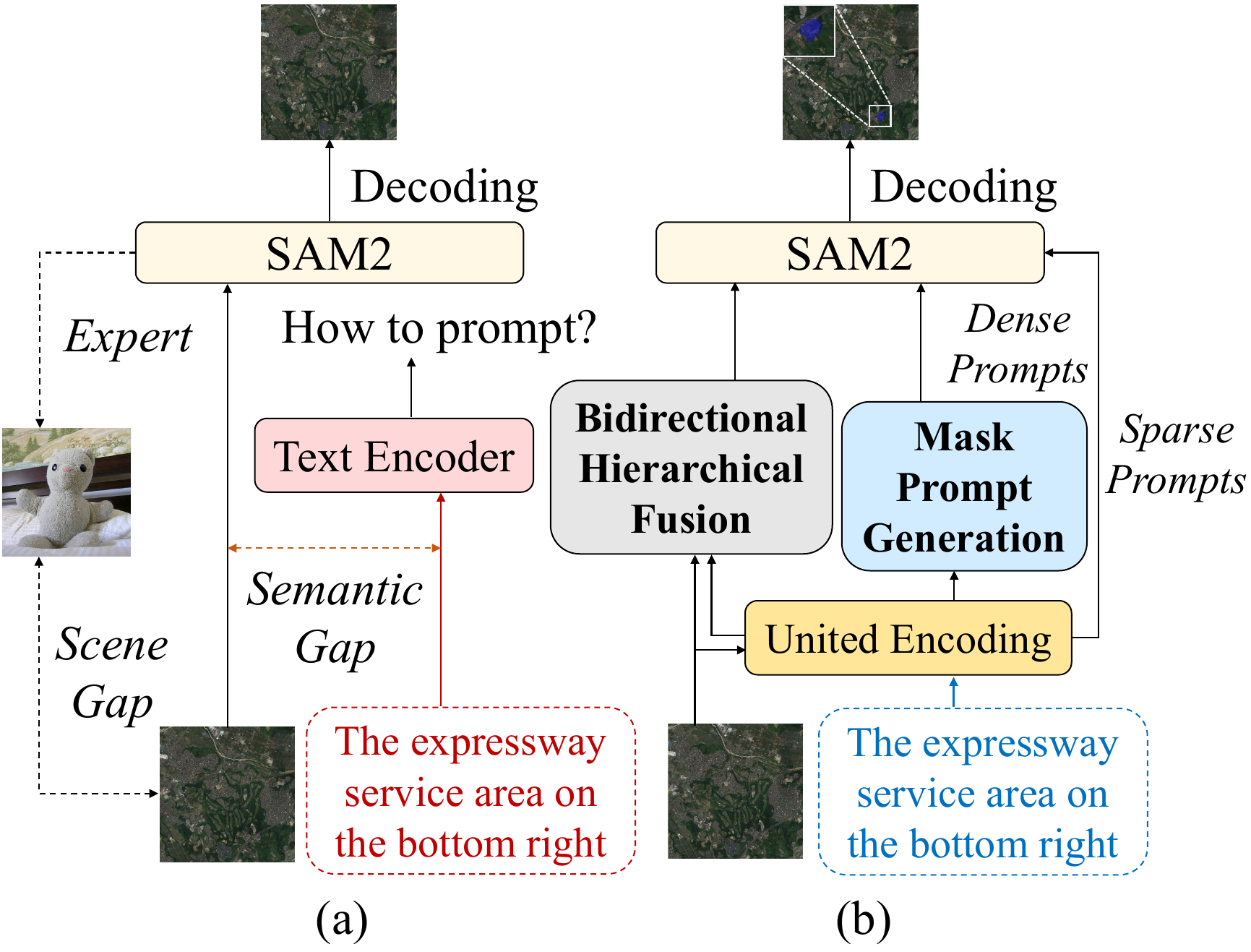}
     \caption{Comparison of two SAM2 adaptations for RRSIS. (a) vanilla SAM2, (b) our RS2-SAM2.}
     \label{fig1}

\end{figure}

Recent advances in the Segment Anything Model (SAM) \cite{kirillov2023segment} and its variants \cite{zhang2023faster, xiong2024efficientsam, ke2024segment} have demonstrated significant improvements in efficiency and accuracy for promptable segmentation tasks in natural images. These models exhibit powerful segmentation capabilities and robust interactive prompting mechanisms. Notably, SAM2 \cite{ravi2024sam} incorporates the hierarchical encoder Hiera \cite{ryali2023hiera}, enhancing its ability to process images with diverse spatial scales. However, despite these advancements, applying SAM2 to RRSIS remains challenging due to the unique complexities of remote sensing images.

\textbf{First}, while SAM2 performs well on natural images, its effectiveness declines in remote sensing scenes due to the low target distinguishability and limited foreground–background contrast. Moreover, SAM2 struggles to exploit spatial information from textual descriptions, as shown in Fig.~\ref{fig1} (a). Although existing studies have attempted to address these issues, significant room for improvement remains. For example, SAM2-Adapter \cite{chen2024sam2} modifies SAM2's encoding process to better adapt to complex scenes. However, this unimodal approach lacks hierarchical information interaction, resulting in an insufficient fine-grained understanding of textual information. Consequently, achieving effective alignment of vision-language information and improving generalization to complex remote sensing scenes remain critical challenges in adapting SAM2 for RRSIS. \textbf{Second}, SAM2’s lack of textual prompt integration limits its ability to generate prompts aligned with textual descriptions.  Previous RRSIS methods, such as RMSIN \cite{liu2024rotated}, rely on independent encoding for visual and textual inputs, but such traditional encoder-decoder structures fail to effectively integrate textual information into SAM2. EVF-SAM \cite{zhang2024evf} mitigates this by jointly encoding both modalities and generating sparse prompts via an MLP, achieving promising results on natural images. However, in remote sensing, sparse prompts alone cannot ensure fine-grained understanding, especially for subtle or indistinct objects. Designing effective text-based prompts to guide decoding therefore remains a key challenge for adapting SAM2 to RRSIS.

To address these challenges, this paper proposes RS2-SAM2, a novel RRSIS framework adapted from SAM2. As illustrated in Fig.~\ref{fig1} (b), our approach focuses on two key aspects:  \textit{(1) adapting SAM2 image features to remote sensing scenes and aligning them with textual features during the image encoding process, and (2) generating dense prompts for precise segmentation.} Specifically, to adapt SAM2 to remote sensing scenes and align remote sensing visual features with textual features, we design a bidirectional hierarchical fusion module, which is embedded both within and after the frozen SAM2 image encoder. Initially, we utilize a union encoder to jointly encode visual and textual inputs, producing semantically aligned visual and textual embeddings as well as multimodal class tokens. Subsequently, our proposed fusion module hierarchically aligns the visually enhanced textual embeddings with adapted remote sensing visual features in the SAM2 encoder.

To equip SAM2 with more precise prompts, we introduce a mask prompt generator. This module integrates jointly encoded visual embeddings and multimodal class tokens to generate a pseudo-mask representing remote sensing target objects. The pseudo-mask is then fed into the SAM2 prompt encoder as a dense prompt, providing pixel-level positional information to enhance the model's segmentation capability.

Experimental results across several RRSIS benchmarks demonstrate the superior performance of our model and the effectiveness of the proposed modules. The primary contributions of this work are outlined as follows:
\begin{itemize}
\item We propose the RS2-SAM2 framework, which adapts the SAM2 model to the RRSIS task by hierarchically aligning the adapted remote sensing visual and textual features, providing pseudo-mask-based dense prompts, and enforcing boundary constraints.

\item We design a bidirectional hierarchical fusion module that adapts the SAM2 encoder to the remote sensing images and aligns the adapted visual features with textual features during the encoding process.

\item We develop a mask prompt generator that leverages multimodal features to produce a pseudo-mask as the dense prompt, offering precise target cues for SAM2.

\end{itemize}

\begin{figure*}[t]
    \centering
     \includegraphics[width=\linewidth]{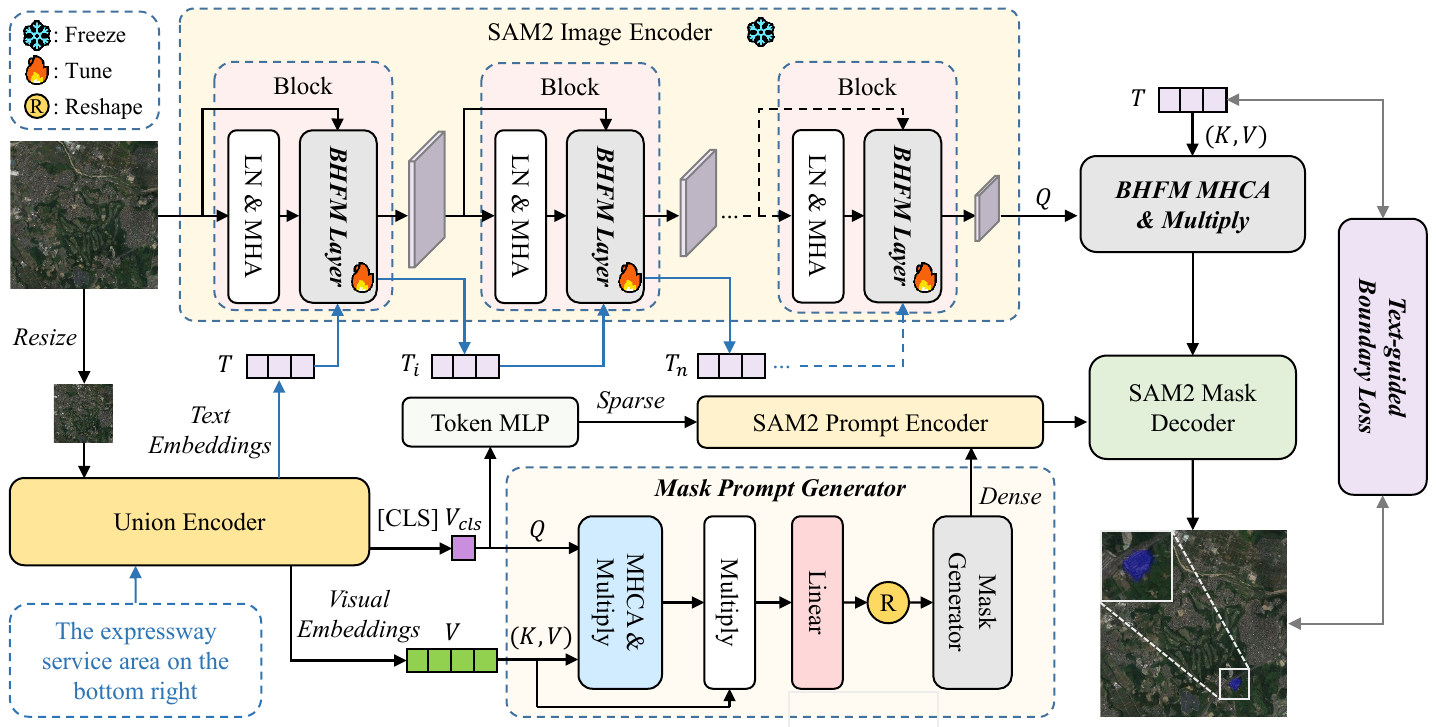}
     \caption{The overview of the proposed RS2-SAM2 framework. It consists of four key components: the union encoder, the bidirectional hierarchical fusion module, the mask prompt generator, and SAM2. The union encoder extracts multimodal representations from the input image and text. The bidirectional hierarchical fusion module enhances image features with textual embeddings. The mask prompt generator produces a prior mask as the dense prompt for SAM2. Finally, SAM2 generates precise masks, while the text-guided boundary loss constrains their boundary accuracy.}
 
     \label{framework}
\end{figure*}

\section{Related Work}

\noindent\textbf{Referring Remote Sensing Image Segmentation.} RRSIS emerges as an important research direction in remote sensing \cite{wang2025sopseg, wang2025pcp}, allowing users to extract geospatial information through natural language queries \cite{wang2025rsood}. Compared to the remote sensing visual grounding (RSVG) \cite{sun2022visual, zhan2023rsvg, lan2024language} task, RRSIS focuses on fine-grained pixel-level analysis rather than region-level identification. However, research in this area remains in its early stages with limited exploration. Yuan et al. \cite{yuan2024rrsis} pioneer this task by introducing the RefsegRS dataset and adapting the LAVT \cite{yang2022lavt} framework for remote sensing. To address challenges like small and fragmented targets easily confused with complex backgrounds, they develop a multi-level feature fusion mechanism incorporating linguistic guidance, significantly enhancing small object detection. Subsequently, Liu et al. \cite{liu2024rotated} build upon the RSVG dataset DIOR-RSVG \cite{zhan2023rsvg} to establish the first large-scale RRSIS dataset, RRSIS-D. They further introduce the RMSIN architecture, which leverages rotational convolution operations to better handle spatial-scale variations and directional complexities in remote sensing images, providing a more robust modeling framework for RRSIS. Building upon RMSIN \cite{liu2024rotated}, Lei et al. propose FIANet \cite{lei2024exploring}, which focuses more on fine-grained vision-language interaction and adaptive understanding of objects at different scales.

\noindent\textbf{Segment Anything Model.} SAM \cite{kirillov2023segment} is an interactive model designed to generate non-semantic segmentation masks based on a variety of input prompts. Leveraging large-scale training data, it demonstrates strong generalization across diverse segmentation tasks. SAM has also been widely utilized in multiple domains, including remote sensing image analysis \cite{wang2024samrs, wang2025tasam, wang2024sampolybuild}, video object tracking \cite{cheng2023segment, yan2024visa}, and medical image segmentation \cite{yue2024surgicalsam, cheng2024unleashing}. To improve its accuracy and efficiency, several optimized versions \cite{zhang2023faster, xiong2024efficientsam, zhong2024convolution} have been developed. More recently, SAM2 \cite{ravi2024sam} has been introduced for video segmentation \cite{rong2025mpg, lan2023learning}, incorporating multi-scale feature encoding to enhance segmentation robustness. Despite these advances, SAM remains limited by its lack of linguistic understanding, preventing it from directly handling referring segmentation tasks that require textual guidance. To address this limitation, recent works \cite{lai2024lisa, xu2023u, rasheed2024glamm} have explored integrating multimodal large language models (MLLMs) to enhance SAM’s ability to process language-based instructions. LISA \cite{lai2024lisa}, for example, fine-tunes LLaVA \cite{liu2024visual} to derive hidden embeddings, thus producing multimodal features for enhanced segmentation. u-LLaVA \cite{xu2023u} extends this paradigm by enabling simultaneous region- and pixel-level segmentation through multi-task learning. With its lightweight pre-fusion strategy, EVF-SAM \cite{zhang2024evf} leverages joint visual-language encoding to strengthen text-driven segmentation prompts, achieving superior segmentation accuracy. However, while these models demonstrate strong performance on natural images, their effectiveness in complex remote sensing scenes \cite{zhang2025uniuir} remains limited, underscoring the need for further refinement and adaptation.

\section{Method}
\subsection{Overview}
The architecture of the proposed RS2-SAM2 framework is depicted in Fig.~\ref{framework}. RS2-SAM2 comprises four essential parts: the union encoder, the bidirectional hierarchical fusion module, the mask prompt generator, and the SAM2 model. Given an input remote sensing image $\mathcal{I}$ and its associated textual description $\mathcal{E}=\left\{e_l\right\}_{l=1}^L$, where $L$ is the number of words, the union encoder processes both modalities to extract a multimodal [CLS] token, visual patch embeddings, and textual embeddings. The SAM2 image encoder, incorporating the bidirectional hierarchical fusion module, further refines the extracted image features by leveraging textual information. Next, the mask prompt generator utilizes the visual patch embeddings and the multimodal [CLS] token to produce a pseudo-mask estimate for the target object. This prior mask, along with the multimodal [CLS] token, serves as a guiding signal for SAM2. Ultimately, the SAM2 decoder synthesizes the extracted image features and generated prompts to produce high-precision segmentation masks.

\subsection{Feature Extraction}

To align visual and textual modalities in RRSIS, this work uses the union BEiT-3 encoder~\cite{wang2023image}, following~\cite{zhang2024evf, yu2024simple}. Each input image $I_u \in \mathbb{R}^{H_u \times W_u \times 3}$ is divided into non-overlapping patches $P_v \in \mathbb{R}^{N_p \times (p^2 \times 3)}$ and then projected to $P_v \in \mathbb{R}^{N_p \times D}$, where $N_p = \frac{H_u \times W_u}{p^2}$ and $D$ is the embedding dimension. A visual class token $V_{\text{cls}} \in \mathbb{R}^{1 \times D}$ and positional embedding $V_{\text{pos}} \in \mathbb{R}^{(N_p + 1) \times D}$ are prepended to obtain:

\[
V_0 = [V_{\text{cls}}, P_v] + V_{\text{pos}}
\]

Text of length $L$ is tokenized using XLM-Roberta~\cite{conneau2019unsupervised} as $T_{\text{seq}}$. A class token and end-of-sequence marker are added, followed by positional embedding: $T_0 = [T_{\text{cls}}, T_{\text{seq}}, T_{\text{eos}}] + T_{\text{pos}}$, where $T_0 \in \mathbb{R}^{N_t \times D}$ and $N_t = L + 2$.

The concatenated multimodal representation $U_{0}$ is constructed as:

\[
U_0 = [V_0; T_0] \in \mathbb{R}^{(N_p + N_t + 1) \times D}
\]

After multimodal fusion and FFNs, the final representation $U$ is obtained, which is decomposed into $V_{\text{cls}} \in \mathbb{R}^{1 \times D}$, $V \in \mathbb{R}^{N_p \times D}$, and $T \in \mathbb{R}^{N_t \times D}$.

The SAM2 image encoder with the proposed bidirectional hierarchical fusion module encodes the input image $I_s \in \mathbb{R}^{H_s \times W_s \times 3}$. It extracts multi-scale features, and the final-layer output $F_n \in \mathbb{R}^{\frac{H_s}{16} \times \frac{W_s}{16} \times C}$ is used for decoding, where $H_s$, $W_s$, and $C$ denote the height, width, and channel number of the SAM2 image encoder, respectively.

\subsection{Bidirectional Hierarchical Fusion Module}
Although SAM2 demonstrates powerful segmentation capabilities for natural images, it struggles to achieve accurate segmentation for remote sensing images with more complex scenes. Inspired by MSA \cite{wu2023medical}, a straightforward approach is to incorporate linear layers into the SAM2 image encoder to enhance its adaptability to remote sensing images. However, for the RRSIS task, integrating textual information during SAM2's image encoding process could make the model more sensitive to referred objects. To address this, we designed a bidirectional hierarchical fusion module and embedded it into SAM2 image encoder, enabling the visual feature of SAM2 to better adapt to remote sensing scenes and the referring text, thereby achieving more precise segmentation.

\begin{figure}[t]
    \centering
     \includegraphics[width=\linewidth]{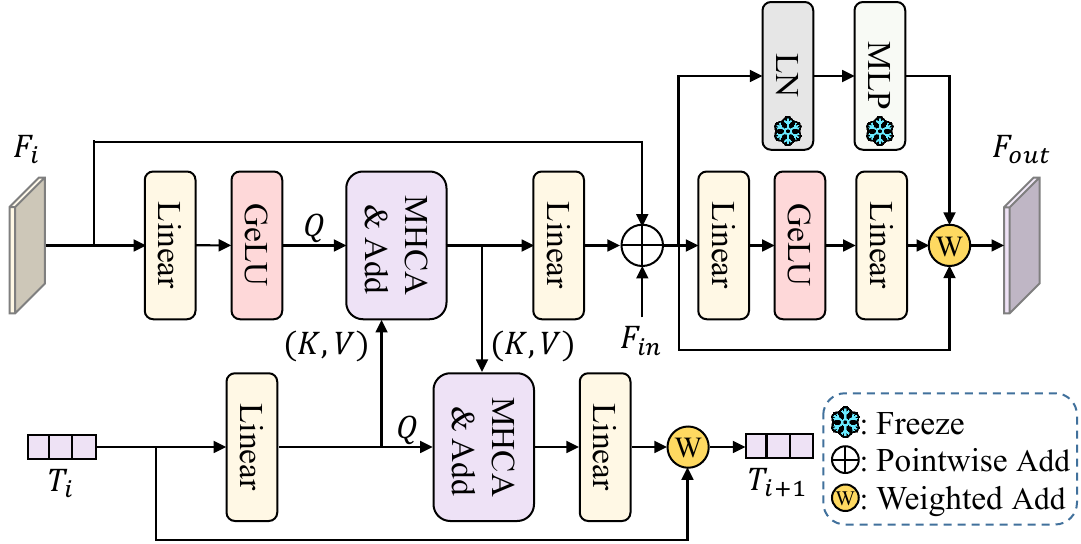}
     \caption{The structure of the bidirectional hierarchical fusion module.}
     \label{BHFM}
\end{figure}

As illustrated in Fig.~\ref{BHFM}, our bidirectional hierarchical fusion module begins with the preprocessed SAM2 image feature $F_i$ of the current layer being projected to a lower dimensionality through a linear layer and an activation function. Simultaneously, the text feature $T_{i}$ of the current layer is also projected to match the dimensionality of the image feature using a linear layer. Subsequently, the image and text features are utilized as query embeddings and key-value pairs, respectively, and undergo cross-attention interaction to capture modality-specific dependencies. The resulting features are then integrated with the pre-interaction representations through element-wise addition. Following this, linear layers are used to restore the dimensionality of both the image and text features. The above process can be expressed by the following equation:
\begin{equation}
\begin{aligned}
    &F_{i}^{'} = \sigma(\text{Linear}(F_{i})), T_{i}^{'} = \text{Linear}(T_{i}),\\
    &F_{i}^{''} = \text{MHCA}(F_{i}^{'}, T_{i}^{'}) + F^{'}_{i},\\
    &T_{i}^{''} = \text{MHCA}(T_{i}^{'}, F_{i}^{'}) + T^{'}_{i},\\
\end{aligned}
\end{equation}
where $\sigma$ denotes the activation function GeLU and MHCA represents the multi-head cross-attention layer. 

To preserve textual integrity during visual enhancement, the text feature is weighted and summed with its pre-interaction representation. Meanwhile, after skip-connection, the visual feature is summed with the original image feature $F_{in}$ of the current layer, then separately processed by the MLP branch and the linear branch, followed by weighted fusion. The entire process is depicted as follows:
\begin{equation}
\begin{aligned}
    T_{i+1} = &(1-\alpha_{t})T_{i} + \alpha_{t}{\text{Linear}(T^{''}_{i})},\\
    F^{'''}_{i} = &F_{in} + \text{Linear}(F_{i}^{''}) + F_{i}, \\
    F_{out} = &F^{'''}_{i} + \text{MLP}(\text{LN}(F^{'''}_{i}))+ \\
    &\alpha_{i}\text{Linear}(\sigma(\text{Linear}(F^{'''}_{i}))),\\ 
\end{aligned}
\end{equation}
here, $\alpha_{t}$ represents the text weighting coefficient, $F_{out}$ denotes the image feature output to the next layer, and $\alpha_{i}$ represents the image weighting coefficient.

After feature encoding, the original text feature $T$ encoded by the union encoder is used to further guide the visual feature $F$ at the high-level. Specifically, the visual feature acts as query and interacts with the text feature as key-value pairs through multi-head cross-attention. The resulting feature is then element-wise multiplied with the visual feature before interaction, yielding the text-guided hierarchical feature $F_{en}$, which is subsequently fed into the SAM2 decoder for accurate decoding.

\subsection{Mask Prompt Generator}

Although the SAM2 image feature has been hierarchically guided by text features using the bidirectional hierarchical fusion module, further providing pixel-level semantic guidance could offer finer-grained textual information. Given the excellent semantic alignment between the visual embeddings $V$ and the textual embeddings $T$ in the joint encoding process, we propose combining the multimodal [CLS] token with visually aligned embeddings that are well-aligned with text. This combination generates a multimodal pseudo-mask prior, which provides pixel-level guidance for the SAM2 decoding process.

As illustrated in Fig.~\ref{framework}, we first employ the multimodal [CLS] token $V_{cls}$ as query and the visual embeddings $V$ as key-value pairs for cross-attention computation. The interaction result is then element-wise multiplied with the multimodal [CLS] token to further align the multimodal token with visual information. Subsequently, the visual embeddings are reshaped into feature maps with dimensions $\frac{H_u}{p} \times \frac{W_u}{p}$, and the multimodal token is passed through a linear layer and broadcasted to match the shape of the visual embeddings. After element-wise multiplication, the result is input into the mask generator, composed of MLP layers, to produce the pseudo-mask $M_{p} \in \mathbb{R}^{\frac{H_u}{p} \times \frac{W_u}{p}}$ that is well-aligned with multimodal information. 

The pseudo-mask $M_{p}$ is subsequently upsampled to dimensions $H_s$ and $W_s$ using linear interpolation, to match the feature size of the SAM2 decoder. This upsampled pseudo-mask is subsequently fed as the dense prompt into the SAM2 prompt encoder.

\subsection{SAM2 Prompt Encoder and Mask Decoder}

The prompt encoder in RS2-SAM2 receives two inputs: a dense prompt $M_p$ from the mask prompt generator and a sparse prompt $V_{\text{cls}}$ from the union encoder. Following EVF-SAM \cite{zhang2024evf}, $V_{\text{cls}}$ is processed via a token MLP and combined with zero-initialized sparse embeddings. To ensure compatibility, the pixel-level prompt $M_{p}$ is adjusted to the spatial dimensions of SAM2 features before being input as the dense spatial information into the mask decoder.

The SAM2 mask decoder leverages both sparse object-level and dense pixel-level prompts: the sparse forms queries with object tokens, while the latter, as a mask prior, is added to visual features to enable direct high-resolution decoding.

\subsection{Training Loss}

RS2-SAM2 utilizes a comprehensive loss function akin to that proposed in \cite{liu2024rotated} to regulate the predicted mask, formulated as follows:
\begin{equation}
\mathcal{L} = \lambda_{ce} \mathcal{L}_{ce} + \lambda_{dice} \mathcal{L}_{dice} + \lambda_{tbl} \mathcal{L}_{tbl},
\end{equation}
where $\mathcal{L}_{ce}$ corresponds to the cross-entropy loss, $\mathcal{L}_{dice}$ denotes the DICE loss \cite{milletari2016v}, and $\mathcal{L}_{tbl}$ signifies the text-guided boundary loss, a loss function we designed that uses text-weighted constraints to predict mask boundaries.

\noindent\textbf{Text-guided Boundary Loss.} In the RRSIS task, unlike natural images, the target objects often exhibit low visual distinguishability from the background, making the boundaries of remote sensing targets less distinct. To address this, we introduce a text-guided boundary loss function $L_{tbl}$ to constrain the predicted mask boundaries. Specifically, we first compute the difference between each pixel value and its neighboring pixel value in both horizontal and vertical directions to obtain the gradient, which serves as an indicator for boundary detection. Then, we abstract the text embeddings $T$ into sentence embeddings $T_{s}$, reduce their dimensionality to a single scalar through a linear layer, and expand it to match the size of the mask as a text-guided weighting factor for the boundary gradient. Finally, we measure the boundary similarity between the predicted mask and the ground truth mask under the guidance of the text weights by MSE loss: 
\begin{equation}
\begin{aligned}
&\nabla_{pre} = Absd_{h}(M_{pre}) +  Absd_{v}(M_{pre})\\
&\nabla_{gt} = Absd_{h}(M_{gt}) +  Absd_{v}(M_{gt})\\
&\mathcal{L}_{tbl} = \frac{1}{N}\sum_{i=1}^{N}(\text{Linear}(T_s)({\nabla_{pre}}_{i} - {\nabla_{gt}}_{i}))^2 \\
\end{aligned}
\end{equation}
where $Absd_{h}$ and $Absd_{v}$ represent the absolute differences between adjacent pixels in the horizontal and vertical directions, respectively, while $\nabla$ denotes the gradient and $N$ is the number of all pixels.

\section{Experiments}
\subsection{Datasets and Metrics}

\noindent\textbf{Datasets.} The experiments are conducted on two key RRSIS datasets: RefSegRS \cite{yuan2024rrsis} and RRSIS-D \cite{liu2024rotated}. RefSegRS is the first remote sensing referring segmentation dataset, containing 2172 images for training, 413 for validation, and 1817 for testing, each with a resolution of 512×512. RRSIS-D is a widely used large-scale dataset in the RRSIS field, including 12181 training samples, 1740 validation samples, and 3481 test samples, with each image sized at 800×800.

\noindent\textbf{Evaluation Metrics.} In accordance with the evaluation protocol established in \cite{liu2024rotated}, we assess our model using several metrics, including precision at various IoU thresholds (Pr@0.5 to Pr@0.9), mean Intersection-over-Union (mIoU), and overall Intersection-over-Union (oIoU). These metrics are computed on the validation and test sets of both RefSegRS and RRSIS-D datasets. 

\subsection{Implementation Details}
\noindent\textbf{Model Settings.} We initialize the key modules of SAM2 and the union encoder using pre-trained weights from SAM2-Hiera-Large \cite{ravi2024sam} and BEiT-3-Large \cite{wang2023image}. For feature extraction, each image is resized to $1024 \times 1024$ and $224 \times 224$, which are then fed into the SAM2 image encoder with an output dimension of $C = 256$ and the union encoder with an output dimension of $D = 1024$. Unlike SAM2’s full capabilities, we do not employ its memory mechanism, focusing exclusively on image processing. In the bidirectional hierarchical fusion module, we set the text weight coefficient $\alpha_{t}$ to 0.2 and the image weight coefficient $\alpha_{i}$ to 0.5.

\noindent\textbf{Training Details.} Experiments are conducted on 8 NVIDIA GeForce RTX 4090 GPUs. The experiment adopts a setup similar to \cite{lei2024exploring}, with training conducted for 60 epochs on the RefSegRS \cite{yuan2024rrsis} dataset and 40 epochs on the RRSIS-D \cite{liu2024rotated} dataset. We use the AdamW optimizer \cite{loshchilov2018decoupled} with a unified batch size of 1. The learning rates for models on RefSegBS and RRSIS-D dataset are set to 5e-5 and 1e-5, respectively. To facilitate the adaptation of our designed modules to remote sensing data, the learning rates for bidirectional hierarchical fusion module and mask prompt generator are assigned as 1e-4 on RefSegRS dataset and 5e-5 on RRSIS-D dataset. All learning rates are reduced to 0.1 times their original values during the last 10 epochs. The weighting coefficients for various loss functions are defined as follows: $\lambda_{dice} = 0.1$, $\lambda_{ce} = 1$, and $\lambda_{tbl} = 0.2$.

\begin{table*}[t]
\centering

\setlength{\tabcolsep}{1mm}
\small
    \begin{center}
        \begin{tabular}{l c c c c c c c c c c c c c c}
    \toprule
    \multirow{2}{*}{Method} & \multicolumn{2}{c}{Pr@0.5} & \multicolumn{2}{c}{Pr@0.6} & \multicolumn{2}{c}{Pr@0.7} & \multicolumn{2}{c}{Pr@0.8} & \multicolumn{2}{c}{Pr@0.9} & \multicolumn{2}{c}{oIoU} & \multicolumn{2}{c}{mIoU} \\
    \cline{2-15}
      & \small Val &\small Test & \small Val &  \small Test & \small Val &  \small Test & \small Val &  \small Test & \small Val &  \small Test & \small Val &  \small Test & \small Val &  \small Test \\
    \midrule
        BRINet \cite{hu2020bi} & 36.86 & 20.72 & 35.53 & 14.26 & 19.93 & 9.87 & 10.66 & 2.98 & 2.84 & 1.14 & 61.59 & 58.22 & 38.73 & 31.51 \\
        LSCM \cite{hui2020linguistic} & 56.82 & 31.54 & 41.24 & 20.41 & 21.85 & 9.51 & 12.11 & 5.29 & 2.51 & 0.84 & 62.82 & 61.27 & 40.59 & 35.54 \\
        CMPC \cite{huang2020referring} &  46.09&  32.36&  26.45&  14.14&  12.76&  6.55&  7.42&  1.76&  1.39&  0.22&  63.55&  55.39&  42.08&  40.63\\
        CMSA \cite{ye2019cross} & 39.24 & 28.07& 38.44 & 20.25& 20.39 & 12.71& 11.79 & 5.61& 1.52 & 0.83& 65.84& 64.53& 43.62 & 41.47\\
        RRN \cite{li2018referring} &55.43 & 30.26 & 42.98 & 23.01 & 23.11 & 14.87 & 13.72 & 7.17 & 2.64 & 0.98 & 69.24 & 65.06 & 50.81 & 41.88 \\
         EVF-SAM \cite{zhang2024evf} & 57.77 & 35.17&37.59& 22.34& 16.24 & 9.36&4.87&2.86 &	1.86&0.39 &59.61 &55.51 & 46.98 & 36.64 \\ 
        CMPC+ \cite{liu2021cross} &56.84&  49.19&  37.59&  28.31&  20.42&  15.31&  10.67&  8.12&  2.78&  2.55&  70.62&  66.53&  47.13&  43.65  \\
        CARIS \cite{liu2023caris} & 68.45 & 45.40 & 47.10 & 27.19 & 25.52 & 15.08 & 14.62 & 8.87 & 3.71 & 1.98 & 75.79 & 69.74 & 54.30 & 42.66 \\
        CRIS \cite{wang2022cris} &53.13 & 35.77 & 36.19 & 24.11 & 24.36 & 14.36 & 11.83 & 6.38 & 2.55 & 1.21 & 72.14 & 65.87 & 53.74 & 43.26 \\
        LAVT \cite{yang2022lavt} & 80.97 & 51.84 & 58.70 & 30.27 & 31.09 & 17.34 & 15.55 & 9.52 & 4.64 & 2.09 & 78.50 & 71.86 & 61.53 & 47.40 \\
        RIS-DMMI \cite{hu2023beyond} &86.17 & 63.89 & 74.71 & 44.30 & 38.05 & 19.81 & 18.10 & 6.49 & 3.25 & 1.00 & 74.02 & 68.58 & 65.72 & 52.15 \\
        LGCE \cite{yuan2024rrsis} &90.72&  73.75&  86.31&  61.14&  71.93&  39.46&  \underline{32.95}&  16.02&  \underline{10.21}&  5.45&  \underline{83.56}&  76.81&  72.51&  59.96\\
        RMSIN \cite{liu2024rotated} &\underline{93.97}&  79.20&  \underline{89.33}&  65.99&  \underline{74.25}&  42.98&  29.70&  16.51&  7.89&  3.25&  82.41&  75.72&  \underline{73.84}&  62.58\\
        FIANet \cite{lei2024exploring} & - &\underline{84.09} & - & \underline{77.05} & - &\underline{61.86} & - &\underline{33.41} & - & \underline{7.10} & - & \underline{78.32} &- & \underline{68.67} \\

        \midrule
        RS2-SAM2 & \textbf{95.36} & \textbf{84.31}& \textbf{94.90} & \textbf{79.42}& \textbf{92.58} & \textbf{70.89}& \textbf{83.76} &\textbf{55.70}& \textbf{36.66} & \textbf{21.19}& \textbf{88.03} & \textbf{80.87}& \textbf{85.21} &\textbf{73.90}\\
        \bottomrule
    \end{tabular}
    \end{center}
\caption{Comparison with state-of-the-art methods on the RefSegRS dataset. The top-performing results are presented in bold, while the second-best results are underlined. Our model achieves the best performance across all metrics.}
    \label{RefSegRS}
\end{table*}

\begin{table*}[t]
\setlength{\tabcolsep}{1mm}
\centering
\small
    \begin{center}
        \begin{tabular}{l c c c c c c c c c c c c c c}
    \toprule
    \multirow{2}{*}{Method} & \multicolumn{2}{c}{Pr@0.5} & \multicolumn{2}{c}{Pr@0.6} & \multicolumn{2}{c}{Pr@0.7} & \multicolumn{2}{c}{Pr@0.8} & \multicolumn{2}{c}{Pr@0.9} & \multicolumn{2}{c}{oIoU} & \multicolumn{2}{c}{mIoU} \\
    \cline{2-15}
     & \small Val &\small Test & \small Val &  \small Test & \small Val &  \small Test & \small Val &  \small Test & \small Val &  \small Test & \small Val &  \small Test & \small Val &  \small Test \\
    \midrule
    RRN \cite{li2018referring}& 51.09 &51.07& 42.47 & 42.11& 33.04 & 32.77& 20.80 & 21.57& 6.14 & 6.37& 66.53 & 66.43& 46.06 & 45.64\\
    CMSA \cite{ye2019cross} & 55.68 & 55.32& 48.04 & 46.45& 38.27 & 37.43& 26.55 & 25.39& 9.02 & 8.15& 69.68 & 69.39& 48.85& 48.54\\
    LSCM \cite{hui2020linguistic} & 57.12 & 56.02& 48.04 & 46.25& 37.87 & 37.70& 26.37 & 25.28& 7.93 & 8.27& 69.28 & 69.05& 50.36& 49.92\\
    CMPC~\cite{huang2020referring} &57.93 & 55.83& 48.85 & 47.40& 38.50 & 36.94& 25.28 & 25.45& 9.31 & 9.19& 70.15 & 69.22& 50.41 & 49.24\\
    BRINet \cite{hu2020bi}& 58.79 & 56.90& 49.54 & 48.77& 39.65 &39.12& 28.21 & 27.03& 9.19 & 8.73& 70.73 &69.88& 51.14 &49.65\\
    CMPC$+$ \cite{liu2021cross} & 59.19 & 57.65& 49.36 & 47.51& 38.67 &36.97& 25.91 & 24.33& 8.16 & 7.78& 70.14 &68.64& 51.41 &50.24\\
    LGCE \cite{yuan2024rrsis}& 68.10 & 67.65& 60.52 & 61.53& 52.24 & 51.45& 42.24 & 39.62& 23.85 & 23.33& 76.68 & 76.34& 60.16 & 59.37\\
    RIS-DMMI \cite{hu2023beyond} & 70.40& 68.74& 63.05& 60.96 &54.14 &50.33& 41.95 &38.38 &23.85 &21.63 &77.01 &76.20 &60.72& 60.12 \\
    LAVT \cite{yang2022lavt}& 69.54 & 69.52& 63.51 & 63.63& 53.16 & 53.29& 43.97 & 41.60& 24.25 & 24.94& 77.59 & 77.19& 61.46 & 61.04\\
    EVF-SAM \cite{zhang2024evf}& 73.51&72.16&67.87&	66.50&	\underline{58.33}&	\underline{56.59}&	\underline{46.15}&\underline{43.92}&\underline{25.92} &\underline{25.48} &76.32& 76.77&64.03	&62.75\\
    FIANet \cite{lei2024exploring}  & - & \underline{74.46} & - &66.96 & - &56.31 & -& 42.83 & - & 24.13 & - &76.91 & - &64.01 \\
    RMSIN \cite{liu2024rotated} & \underline{74.66} & 74.26 & \underline{68.22} & \underline{67.25}& 57.41 & 55.93 & 45.29 & 42.55& 24.43 & 24.53& \underline{78.27} & \underline{77.79}& \underline{65.10} &\underline{64.20}\\
    \midrule
    RS2-SAM2 & \textbf{79.25} & \textbf{77.56}& \textbf{74.08} & \textbf{72.34}& \textbf{63.85} & \textbf{61.76}& \textbf{50.57} &\textbf{47.92}& \textbf{30.40} & \textbf{29.73}& \textbf{80.16} & \textbf{78.99}& \textbf{68.81} &\textbf{66.72}\\
    \bottomrule
  \end{tabular}
    \end{center}
\caption{Comparison with state-of-the-art methods on the RRSIS-D dataset. The top-performing results are presented in bold, while the second-best results are underlined. Our model achieves the best performance across all metrics.}
    \label{RRSIS-D}
\end{table*}

\subsection{Comparison with State-of-the-Art Methods}
\noindent\textbf{RefSegRS dataset.} We conduct a comprehensive comparison between our method, RS2-SAM2, and several state-of-the-art approaches on the RefSegRS \cite{yuan2024rrsis} dataset. The results of the comparison are reported in Tab.~\ref{RefSegRS}. It can be observed that our RS2-SAM2 achieves 88.03\% oIoU and 85.21\% mIoU on the validation set, surpassing the previous best method, RMSIN, by 5.62\% in oIoU and 11.37\% in mIoU. On the test set, it reaches 80.87\% oIoU and 73.90\% mIoU, outperforming FIANet \cite{lei2024exploring} by 2.55\% and 5.23\%, respectively. Moreover, the Pr metric exhibits significant improvements across all thresholds, particularly at Pr@0.7, Pr@0.8, and Pr@0.9, demonstrating the model’s strong multimodal segmentation capabilities.

\noindent\textbf{RRSIS-D dataset.} We also conduct comparative experiments between our RS2-SAM2 and existing methods such as LGCE \cite{yuan2024rrsis}, LAVT \cite{yang2022lavt}, EVF-SAM \cite{zhang2024evf} and RMSIN \cite{liu2024rotated}, etc., on the RRSIS-D \cite{liu2024rotated} dataset, with results documented in Tab.~\ref{RRSIS-D}. On this dataset, our method achieves the best performance, surpassing the current state-of-the-art method RMSIN on the validation set by 4.59\% in Pr@0.5, 5.86\% in Pr@0.6, 6.44\% in Pr@0.7, 5.28\% in Pr@0.8, 5.97\% in Pr@0.9, 1.89\% in oIoU, and 3.71\% in mIoU. It also significantly outperformed existing methods on the test set, demonstrating its strong capability in remote sensing object segmentation.

\begin{figure*}[t]
\begin{center}
   \includegraphics[width=0.95\linewidth]{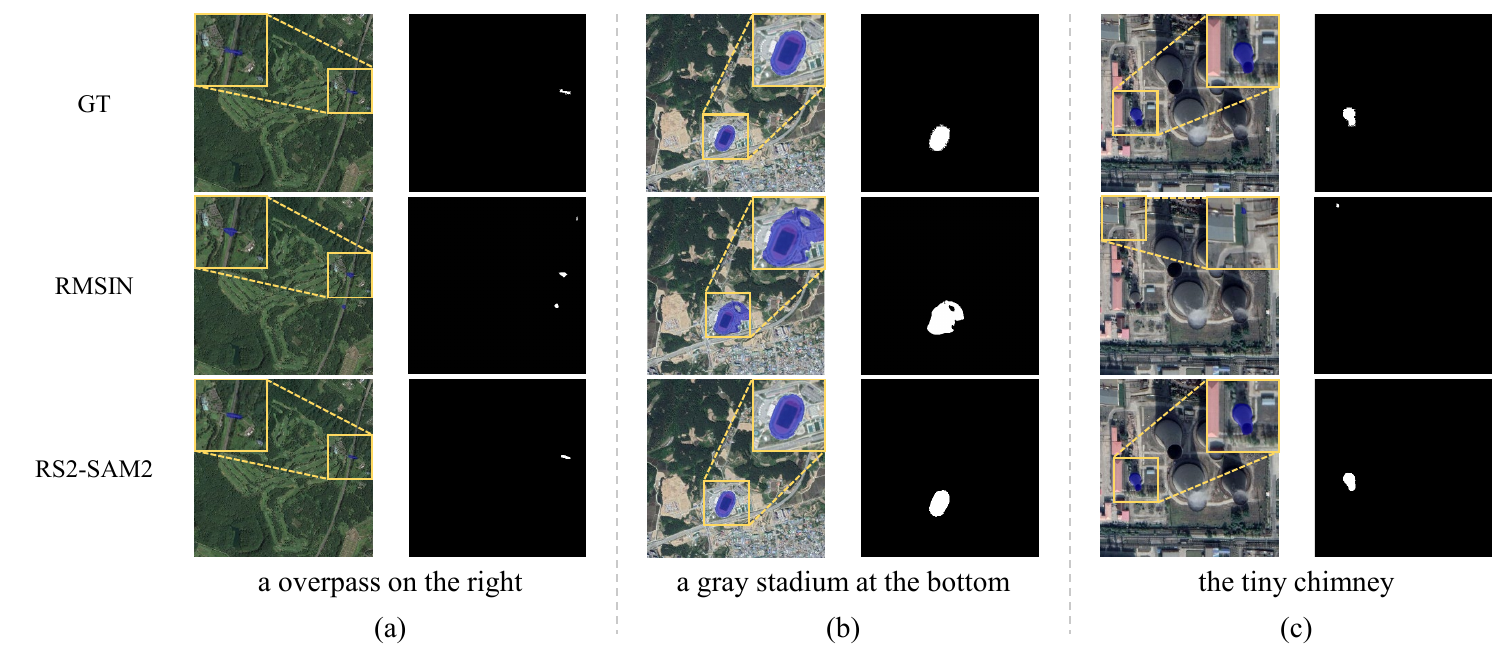}
\end{center}
\caption{Visualization result on RRSIS-D. Compared to RMSIN \cite{liu2024rotated}, RS2-SAM2 demonstrates superior capability in handling local details and boundary regions.}
\label{vis_res}
\end{figure*}

Fig.~\ref{vis_res} illustrates the visual comparison between our model and RMSIN \cite{liu2024rotated} on the RRSIS-D dataset. The results clearly indicate that RS2-SAM2 consistently outperforms RMSIN, particularly in terms of accurate target localization, reliable mask prediction, and boundary precision.

\subsection{Model Analysis}
In this section, we perform extensive ablation studies to analyze the impact of the essential components in our RS2-SAM2 framework, as well as the effects of different model configurations. The experiments are conducted on the test dataset of RefSegRS \cite{yuan2024rrsis}.

\noindent\textbf{Components Analysis.} To investigate the impact of key components in our model, we first construct a baseline model consisting solely of SAM2 and the union encoder. As shown in Tab.~\ref{components}, when the text-guided boundary loss ($\mathcal{L}_{tbl}$) is incorporated during training, the model achieves 38.63\% mIoU and 57.36\% oIoU, surpassing the baseline by 1.99\% and 1.85\%, respectively. Subsequently, on top of the model with $\mathcal{L}_{tbl}$, we add the mask prompt generator (MPG), which results in the RS2-SAM2 achieving 60.20\% mIoU and 70.89\% oIoU, an improvement of 21.57\% and 13.53\% over the previous model. When the bidirectional hierarchical fusion module (BHFM) is added to the baseline model with $\mathcal{L}_{tbl}$, the model shows a further increase of 30.08\% in mIoU and 21.00\% in oIoU, further validating the superiority of this module. Finally, when all components are integrated, our RS2-SAM2 achieves the best performance, with a mIoU of 73.90\% and an oIoU of 80.87\%.

\begin{table}

\setlength{\tabcolsep}{0.23mm}
\centering
\small
    \begin{center}
        \begin{tabular}{l c c c c c c c c}

\toprule
Method & $\mathcal{L}_{tbl}$ & MPG & BHFM & Pr@0.5 & Pr@0.7 & Pr@0.9 & mIoU & oIoU \\
\hline

Baseline & & & &35.17 & 9.36 & 0.39 & 36.64&55.51\\ 
RS2-SAM2 &\ding{51} & & & 39.79 & 12.49 & 0.39 &38.63&57.36\\
RS2-SAM2 &\ding{51} & \ding{51} & &71.00 &43.42& 3.96 &60.20 &70.89\\
RS2-SAM2 & \ding{51} & & \ding{51}&81.89 & 61.86 & 7.10  &68.71&78.36\\
RS2-SAM2 & \ding{51}&\ding{51}& \ding{51}&\textbf{84.31} & \textbf{70.89} & \textbf{21.19}&\textbf{73.90}&\textbf{80.87}\\

\bottomrule

\end{tabular}
    \end{center}
    \caption{Ablation study of different components of RS2-SAM2 on RefSegRS test set.}
    \label{components}
\end{table}

\noindent\textbf{Mask Prompt Generator.} In this section, we explore different interaction forms within the mask prompt generator (MPG), with results presented in Tab.~\ref{ablation}. When the multi-head cross-attention (MHCA) between the multimodal token $V_{cls}$ and visual embeddings $V$ is omitted, the performance of RS2-SAM2 drops by 2.31\% in mIoU and 0.98\% in oIoU. These findings underscore the importance of further strengthening the semantic connections between the multimodal token and visual embeddings during the multimodal mask prompt generation process.

\noindent\textbf{Bidirectional Hierarchical Fusion Module.} The effect of different BHFM configurations is also investigated. We first assess how the BHFM structure in the encoder influences performance. A simple variant uses a linear adapter-like layer without text interaction (``Linear"). Another uses text to enhance visual features without feedback (``Uni"). The third performs bidirectional enhancement between text and vision (``Bi"). As shown in Tab.~\ref{ablation}, the layer-by-layer bidirectional enhancement proves more effective in infusing linguistic cues into visual features, enabling progressive textual refinement from low to high levels and resulting in more precise remote sensing segmentation.

\begin{table}
\setlength{\tabcolsep}{0.5mm}
\centering
\small
    \begin{center}
        \begin{tabular}{l c c c c c c}

\toprule

{Method} & {Settings} &Pr@0.5 & Pr@0.7 & Pr@0.9 & mIoU & oIoU\\

\hline
\multicolumn{5}{l}{\textbf{Interaction Form of MPG}} \\
\hline
RS2-SAM2& w/o MHCA & 83.71 & 67.64 & 16.79&71.59&79.89 \\
RS2-SAM2 & w MHCA & \textbf{84.31} & \textbf{70.89} & \textbf{21.19} & \textbf{73.90} & \textbf{80.87} \\ 
\hline
\multicolumn{5}{l}{\textbf{Structure of BHFM layer}} \\
\hline
RS2-SAM2 & Linear &81.01&61.42&6.77 & 68.19&77.39\\
RS2-SAM2 & Uni & 81.23&63.84&14.14&70.10 &78.93  \\
RS2-SAM2 & Bi &\textbf{84.31} & \textbf{70.89} & \textbf{21.19} & \textbf{73.90} & \textbf{80.87}\\ 
\hline
\multicolumn{5}{l}{\textbf{Components of BHFM}} \\
\hline
RS2-SAM2 & w/o BC & 79.97&58.17&11.61 & 67.61&77.54\\
RS2-SAM2 & w/o BL & 72.21&43.48&4.79&61.08&72.33\\
RS2-SAM2 & w BC\&BL & \textbf{84.31} & \textbf{70.89} & \textbf{21.19} & \textbf{73.90} & \textbf{80.87}\\
\bottomrule

\end{tabular}
    \end{center}
    \caption{Model analysis of different settings in RS2-SAM2.}
    \label{ablation}

\end{table}

Additionally, we investigate the impact of BHFM components on segmentation performance by setting up experiments where the BHFM cross-attention (BC) after encoding and the BHFM layer (BL) without the encoder are removed. The results in Tab.~\ref{ablation} indicate that both interactions during and after encoding are essential. The hierarchical interaction combining both can help the model understand text features from a global to a local perspective, enhancing its pixel-level understanding of text.

\section{Conclusion}

In this paper, we present RS2-SAM2, an advanced end-to-end framework designed to enhance SAM2 for RRSIS. Our approach leverages a union encoder to jointly encode visual and textual features, producing semantically aligned visual-text embeddings and multimodal class tokens. To effectively integrate spatial and textual information, we introduce a bidirectional hierarchical fusion module, which incorporates textual semantics and spatial context both during and after encoding, enabling a hierarchical refinement from global to local levels. Additionally, a mask prompt generator generates multimodal mask as dense prompt, improving the segmentation of visually indistinct objects by providing stronger pixel-level guidance. Extensive experiments on multiple RRSIS benchmarks demonstrate the superiority of RS2-SAM2 over state-of-the-art methods, validating the effectiveness of our proposed modules.

\section*{Acknowledgments}
This work was supported by the National Natural Science Foundation of China under Grant 62431020, the National Key Research and Development Program of China under Grant 2024YFE0111800, and the Fundamental Research Funds for the Central Universities under Grant 2042025kf0030.
\bibliography{main}

\end{document}